\pdfoutput=1
\documentclass[runningheads]{llncs}
\usepackage[T1]{fontenc}
\usepackage{graphicx}
\usepackage{float}
\usepackage{siunitx}
\usepackage{hyperref}
\usepackage{booktabs}
\begin{document}
\title{SoftPINCH: EMG-Driven Soft Exoskeleton Assistance for Finger Flexion and Grasping}
\titlerunning{SoftPINCH}
\author{Nicklas Nikolaj Grønvall$^{1}$, Magnus Malthe Sigsgaard Nielsen$^{1}$, Xiaofeng Xiong$^{1}$*, Saravana Prashanth Murali Babu$^{1}$*}
\authorrunning{Grønvall, N. N. et al.}

\institute{SDU Soft Robotics, The Maersk Mc-Kinney Moller Institute,\\
University of Southern Denmark, Odense, Denmark \\
\email {spmb@mmmi.sdu.dk}, {nicklasgroenvall@gmail.com}}
\maketitle              % typeset the header of the contribution

\begin{abstract}
Surface electromyography (sEMG) provides a non-invasive interface for detecting hand-movement intention and controlling wearable assistive devices. However, reliable EMG-driven hand assistance remains challenging because EMG signals are affected by noise, motion artifacts, electrode placement, muscle fatigue, and inter-subject variability. At the same time, many hand exoskeletons remain mechanically restrictive or bulky, limiting comfort and natural hand motion. This work presents SoftPINCH, an EMG-driven soft wearable exoskeleton for thumb–index finger flexion and pinch grasp assistance. The system combines a tendon-driven soft exoskeleton, fingertip magnetic contact sensing, and neural EMG decoding for intention-based assistance. Surface EMG was recorded from forearm muscles during index and thumb movements, and three subject-independent decoding architectures were evaluated: LSTM, CNN$+$LSTM, and CNN+LSTM with attention. The CNN+LSTM and CNN+LSTM-attention models both achieved 99.4\% LOSO test accuracy, outperforming the standalone LSTM, which reached 97.8\%. However, the attention mechanism did not provide a significant improvement over CNN+LSTM, indicating that CNN-based feature extraction was sufficient for robust EMG representation. The CNN+LSTM model was therefore selected for real-time deployment due to its high accuracy and lower architectural complexity. Functional evaluation showed that active exoskeleton assistance reduced muscular effort during isolated finger flexion and object grasping. During weighted grasping, assistance reduced muscular effort across all tested loads, with a 92.6\% reduction at the highest load. These results demonstrate the potential of SoftPINCH for intuitive, low-effort pinch assistance using real-time EMG-driven soft robotic control.

\end{abstract}

\section{Introduction}
Hand dexterity is essential for independent daily living. Everyday activities such as eating, writing, dressing, tool use, communication, and object manipulation depend on finger motion, grasp formation, and appropriate force control~\cite{intro_Dollar2014}. Impairments caused by stroke, spinal cord injury, neurological disorders, trauma, or age-related weakness can reduce hand mobility and make simple daily tasks difficult to perform~\cite{intro_WearableTech}.
Wearable hand exoskeletons have therefore gained attention as assistive and rehabilitation technologies that can support impaired hand function, restore prehensile movements, and reduce muscular effort during repeated or demanding tasks like grasping~\cite{intro_exo_review}.\\
Among different grasping patterns, pinch grasp is particularly important because it enables fine and precise manipulation~\cite{intro_thumb_evolution}.
The coordination between thumb and index finger allows humans to pick up small objects, button clothes, hold a pen, operate tools, and interact with electronic devices. Classical grasp studies distinguish precision grips from power grips based on the functional role of the hand during object interaction, while modern grasp taxonomies identify thumb-finger and lateral pinch patterns as important grasp types used in everyday manipulation tasks~\cite{intro_7243327,intro_Prehensile_movements}. Thumb position is a key feature of human dexterity ~\cite{intro_thumb_evolution}, and pinch strength has also been associated with functional independence in activities of daily living, particularly in individuals with impaired hand function~\cite{intro_GripNPinch}. For assistive devices, pinch grasp is therefore a relevant target because it requires not only sufficient force generation, but also coordinated thumb-index motion, precise timing, and comfortable interaction with the user.
\begin{figure}[H]
    \centering
    \includegraphics{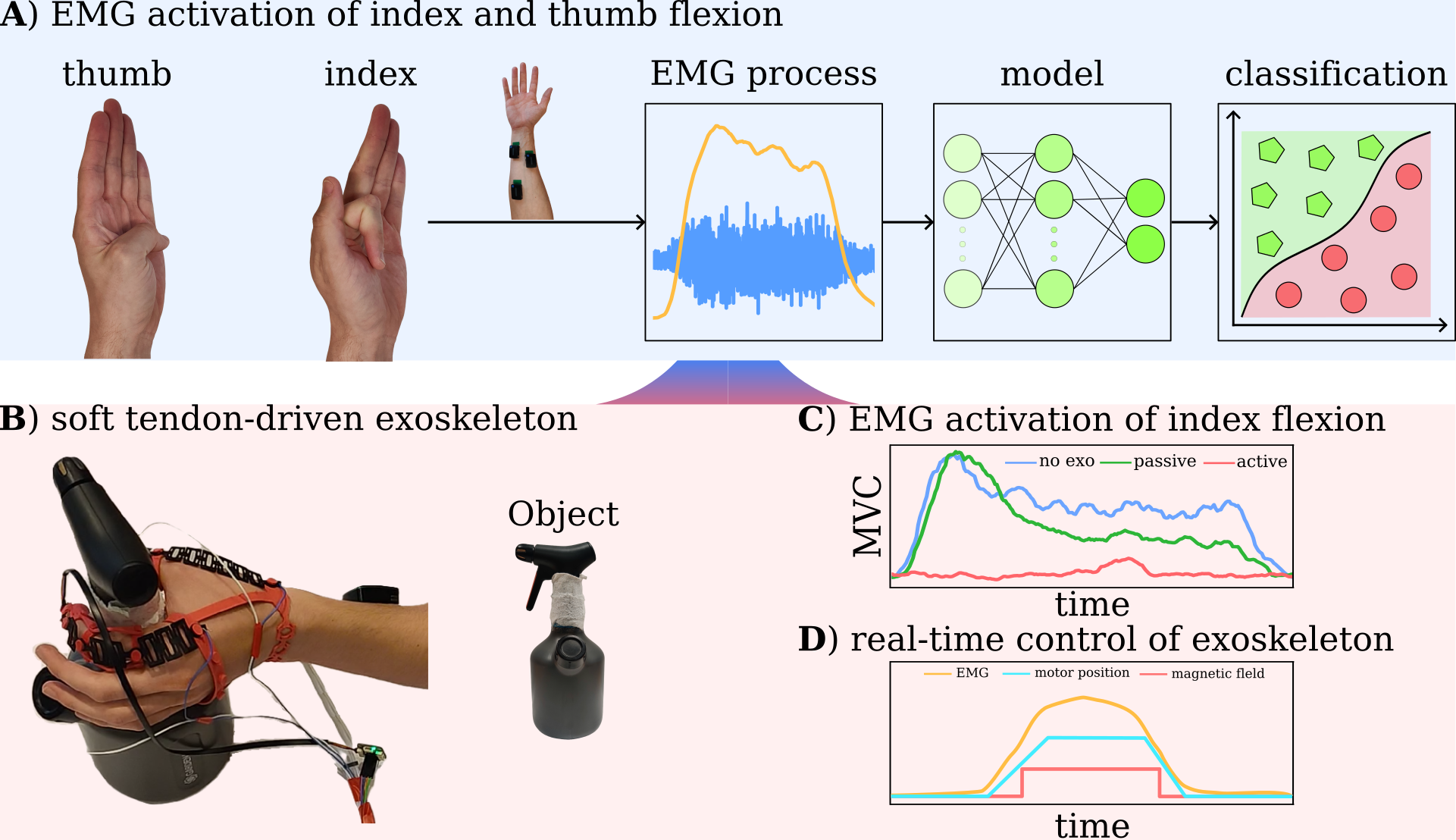}
    \caption{Overview of SoftPINCH.}
    \label{fig:intro_abstract_figure}
\end{figure}

\noindent Hand exoskeletons have evolved from rigid linkage-based systems towards softer and more wearable architectures. Rigid exoskeletons can provide accurate motion guidance and high force transmission and have shown strong potential for rehabilitation and force augmentation~\cite{intro_exo_review,intro_8745535}.
However they often require careful joint alignment, include bulky mechanical structures, and may restrict natural hand motion or reduce comfort during long-term use. Soft robotic hand exoskeletons address some of these limitations by using complaint materials, tendon driven mechanisms, fabric structures, or fluidic actuators that better conform to the hand and allow more natural movement~\cite{intro_soft_glove_art,intro_Soft_device_rehab}. Hybrid rigid\text{-}soft designs have also been explored to combine mechanical support with improved wearability and user intention control~\cite{intro_soft_rigid_art}. Despite these advances, soft hand exoskeletons still face challenges in compact actuation, reliable force transmission, sensing integration, and intuitive control~\cite{intro_Soft_device_rehab,intro_exo_review}. These challenges are especially important for pinch assistance, where the device must remain lightweight and comfortable while coordinating the thumb and index finger during contact\text{-}rich object manipulation~\cite{intro_8745535}.\\
Surface electromyography (sEMG) is a non\text{-}invasive technique in which electrodes are placed on the skin to measure electrical activity produced during muscle contractions.~\cite{intro_EMG_control_art} demonstrated real\text{-}time EMG\text{-}based pattern recognition for stroke patients, strengthen grip requirement through a hand exoskeleton.
However, EMG remains sensitive to electrode placement, cross\text{-}talk, motion artifacts, muscle fatigue, and inter\text{-}subject variability, which might degrade signal accuracy over time.
Among these a computational lag appears in multi\text{-}channel EMG processing, causing real-time application feel slow to the user.
Recurrent neural networks, such as Long Short-Term Memory (LSTM), have shown strong potential for modelling sequential data and capturing long\text{-}term dependencies~\cite{intro_LSTM,intro_LSTM_prediction_art}.
This field was later reshaped by Transformer architectures, where self\text{-}attention replaced recurrent processing to improve sequence modelling and reduce computational cost in translation tasks~\cite{intro_ATTENTION_art}.
For high\text{-}frequency data streams, WaveNet showed that dilated causal convolutions can expand the receptive field and computational more efficient than LSTMs on very long sequences~\cite{intro_wavenet_art}.
Within biorobotics, EMG\text{-}based gesture recognition has demonstrated that CNNs can effectively decode EMG signals, while RNNs remain useful for capturing more complex temporal dynamics~\cite{intro_emg-based_recognition_art}.
However, many of these models remain computationally demanding for long sequences and high\text{-}dimensional representations, leading to inference bottlenecks that can limit real\text{-}time robotic control.\\
Fig.~\ref{fig:intro_abstract_figure} provides an overview of the SoftPINCH system and workflow, an EMG\text{-}driven soft wearable hand exoskeleton for finger flexion and pinch grasp assistance. The system integrates a tendon\text{-}driven thumb\text{-}index exoskeleton, EMG\text{-}based motion decoding, and fingertip sensing for object\text{-}contact feedback. Subject\text{-}independent motion recognition is evaluated using three neural network architectures: LSTM, CNN+LSTM, and CNN+LSTM with attention. The functional effect of the exoskeleton is assessed by comparing EMG\text{-}derived muscular effort across three conditions: no exoskeleton, passive exoskeleton, and active assistance during isolated finger flexion and object grasping.
We hypothesize that adding convolutional feature extraction to an LSTM improves robustness and generalizability of EMG\text{-}based thumb and index finger motion decoding compared with a standalone LSTM. We further hypothesize that the soft exoskeleton does not significantly increase muscular effort when worn passively, while active assistance reduces muscular demand during pinch\text{-}related finger flexion and grasping tasks.

\section{Materials and Methods}

\subsection{Hand Exoskeleton System}\label{sec:hand_exo}
\begin{figure}[H]
    \centering\includegraphics{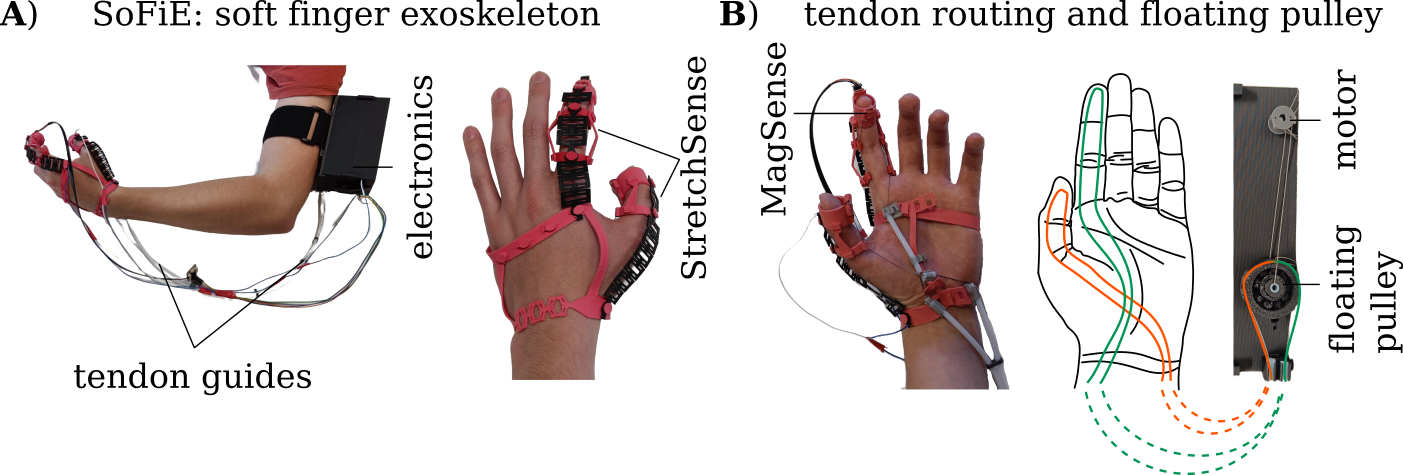}
    \caption{\textbf{A}) SoFiE the soft finger exoskeleton shown in its entirety mounted on a subject's arm with an overview of the main components. \textbf{B}) Picture and illustration of the tendon routing and how the tendon is connected at the floating pulley at the electronics housing.}
    \label{fig:exo}
\end{figure}

For testing and demonstration, we used the SoFiE platform~\cite{SoFiE}, a custom soft finger exoskeleton for thumb-index grasp assistance (Fig.~\ref{fig:exo}\textbf{A}). The system is a modular, 3D-printed, tendon-driven glove that actuates fingers with a single motor made primarily of Colorfabb varioShore TPU. Each finger module includes StretchSense, a conductive compliant spring made of Recreus conductive filaflex TPU that returns the finger to its neutral position and can be used to estimate finger motion through strain-dependent resistance. The index finger includes MagSense, a magnetic fingertip sensor that measures contact-induced changes in the magnet-magnetometer distance as a proxy for touch or grasping force.

\noindent Tendon actuation is routed from the upper-arm electronics housing to the glove through low-friction PTFE tubes and connected to a floating pulley (Fig.~\ref{fig:exo}\textbf{B}). The pulley distributes motor force between the thumb and index finger, allowing one finger to continue moving even if the other is constrained by an object. The electronics are mounted on the upper arm to reduce distal hand mass and include a \SI{7.4}{V} LiPo battery, a Pololu 250:1 micro metal gear motor, an ESP32-based control board, motor driver, and \SI{5}{V} converter.
\subsection{EMG Acquisition and Processing}\label{sec:processing}
\begin{figure}[t]
    \centering
    \includegraphics{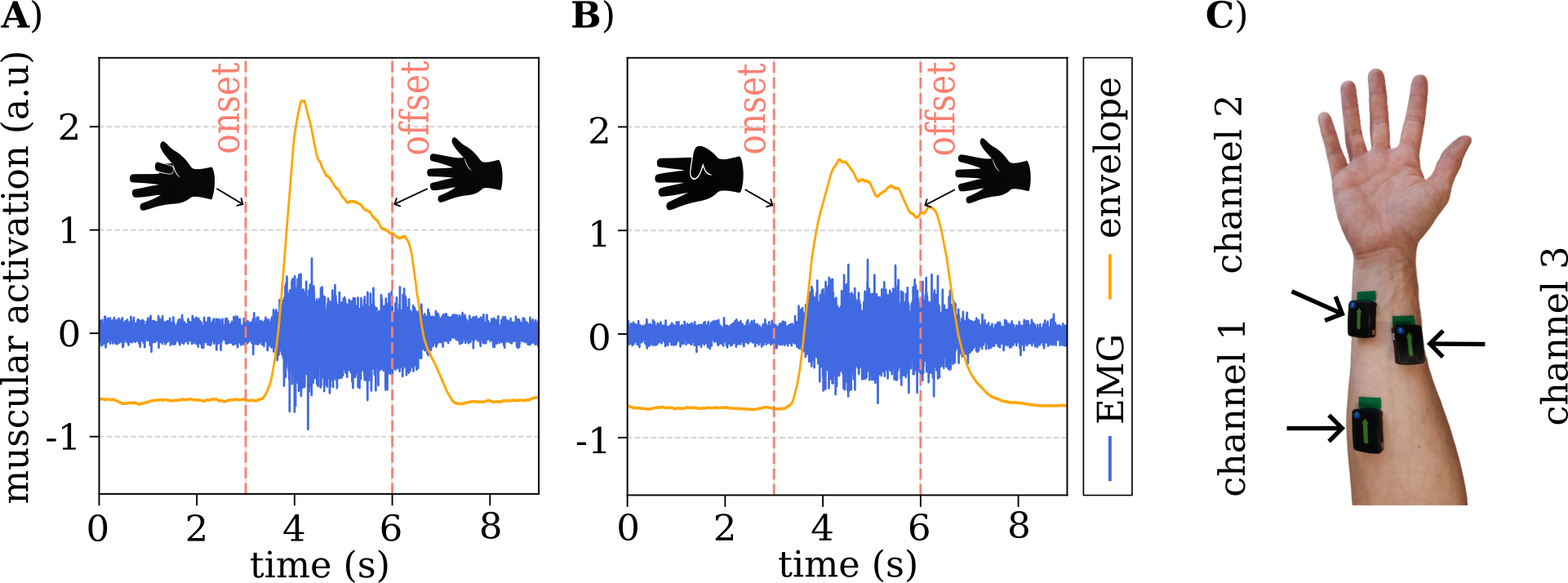}
    \caption{\textbf{A}) shows the raw and processed EMG (envelope) during index finger flexion at channel 2. \textbf{B}) displays thumb flexion at channel 3. \textbf{C}) shows channel locations, channel 1 is placed on the wrist flexor group, channel 2 at the flexor digitorum superficialis, and channel 3 at the flexor pollicis longus.}
    \label{fig:met_EMG_presentation}
\end{figure}

EMG data were acquired using a Trigno wireless biofeedback base station with Trigno Avanti$^{TM}$ sensors, sampled at \SI{2000}{Hz}. Three sensors were placed on the anterior forearm of the participant's dominant hand to distinguish finger flexion movements (Fig.~\ref{fig:met_EMG_presentation}\textbf{C}). Channel 1 targeted the superficial wrist flexor group, channel 2 the flexor digitorum superficialis, and channel 3 the flexor pollicis longus, corresponding to muscles involved in wrist and finger flexion~\cite{BCIMeth_EMG_placement_okwumabua}. Experimentally, these placements showed the strongest responses for ring/middle, index/pinky, and thumb flexion, respectively.
The raw EMG was filtered to suppress motion artifacts, high-frequency noise, and power-line interference. Based on the Trigno user guide and related literature~\cite{BCIMeth_delsysTrignoUserGuide,BCIMeth_RMSandFiltering_Phinyomark,BCIMeth_slidingWindow_Lee}, a 4th-order Butterworth bandpass filter at $20\text{--}450$~\si{Hz} and a 2nd-order \SI{50}{Hz} notch filter were applied. Three seconds of idle recording were added before and after each session and later removed to avoid start-up instability, data loss, and filter edge effects. Because closely located forearm muscles can cause EMG crosstalk~\cite{BCIMeth_hampel_Bhowmik}, and involuntary twitches may introduce impulsive artifacts, a Hampel filter was applied to repair outliers in the signal~\cite{BCIMeth_hampel_Bhowmik}. PSD analysis showed that these artifacts mainly occurred around $180\text{--}220$~\si{Hz}, within the selected EMG frequency range. The Hampel filtered EMG signal was then converted into an RMS envelope using a \SI{250}{ms} sliding window with \SI{25}{ms} increments, corresponding to $90\%$ overlap. Finally, the EMG data were Z-score normalized and segmented into \SI{9}{s} trials.
The original and processed EMG are shown in Fig.~\ref{fig:met_EMG_presentation}\textbf{A} and \textbf{B}

\subsection{Neural Decoding Framework}
\begin{figure}[H]
    \centering
    \includegraphics{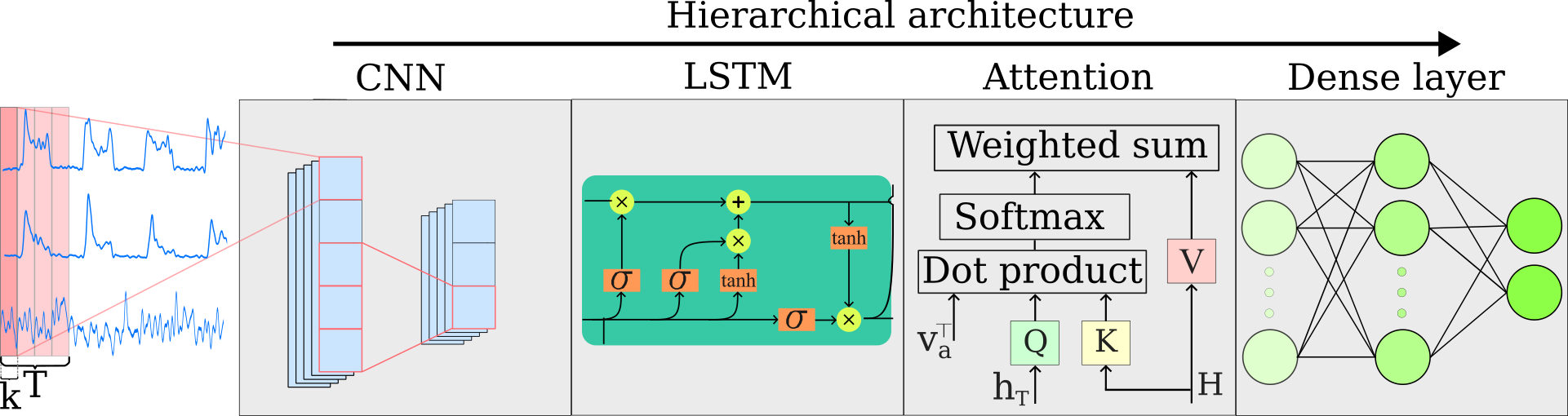}
    \caption{Hierarchical CNN+LSTM+attention architecture. The CNN extracts local temporal features, the LSTM models sequential dependencies, and the attention mechanism weights the most informative time steps before final classification through a dense layer.}
    \label{fig:met_model_architecture}
\end{figure}

The neural decoding framework was developed to classify motor execution from preprocessed EMG signals. Since EMG-based finger-motion decoding is affected by inter-subject variability, this study evaluates three network architectures with increasing complexity to capture temporal and spatial features from the EMG sequence. The overall architecture is shown in Fig.~\ref{fig:met_model_architecture}. The models are referred to as $N_{1}$: LSTM, $N_{2}$: CNN+LSTM, and $N_{3}$: CNN+LSTM with attention.\\
$N_{1}$ uses an LSTM to model temporal dependencies directly from the preprocessed EMG sequence. As illustrated in Fig.~\ref{fig:met_model_architecture}, the LSTM gates control what information is retained, updated, or discarded, allowing relevant long-term sequence information to be preserved.\\
$N_{2}$ extends $N_{1}$ by adding two 1D CNN in sequence before the LSTM. The CNN extracts local temporal and cross-channel features using learned filters, and the resulting feature map are passed to the LSTM to model temporal dependencies over time \\
$N_{3}$ further extends $N_{2}$ by adding an attention mechanism inspired by attention pooling without a decoder~\cite{BCIMeth_attentionPooling_santos}. The attention block assigns weights to the LSTM hidden states, allowing the model to emphasize the most informative time steps before classification.

\noindent Together, these three architectures allow us to evaluate whether increasing model complexity improves subject-independent EMG decoding, and whether attention provides additional benefit beyond the CNN+LSTM representation.

\subsection{Training Strategy}
Hyperparameter optimization was performed using Bayesian optimization through the SHERPA Python library.
The search algorithm explores the hyperparameter space and learns which configurations provide the lowest validation loss.
The LSTM utilizes hidden units, bidirectional structure, and layers, whereas the CNN includes kernel size and filters. Additional hyperparameters are dedicated to the optimizer (AdamW) and dense layers.
For each architecture, 100 hyperparameter configurations were evaluated.
Training ran for up to 250 epochs with a 25-epoch early stopping patience.
Subject-independent classification used a leave-one-subject-out (LOSO) strategy, partitioning the 17 participants into a 14/2/1 train-validation-test split.
Performance was evaluated via 8-fold cross-validation.
Each fold designated two random participants for validation and one unseen participant for testing.
Data leakage was prevented by ensuring individual subjects, remained strictly within a single split.
Models were trained using cross-entropy loss and the AdamW optimizer, selected for its momentum and weight decay properties to improve convergence and generalization~\cite{BCIMeth_adamW_Ilya}.

\subsection{Experimental Setup}\label{sec:experimental_setup}

\begin{figure}[H]
    \centering
    \includegraphics{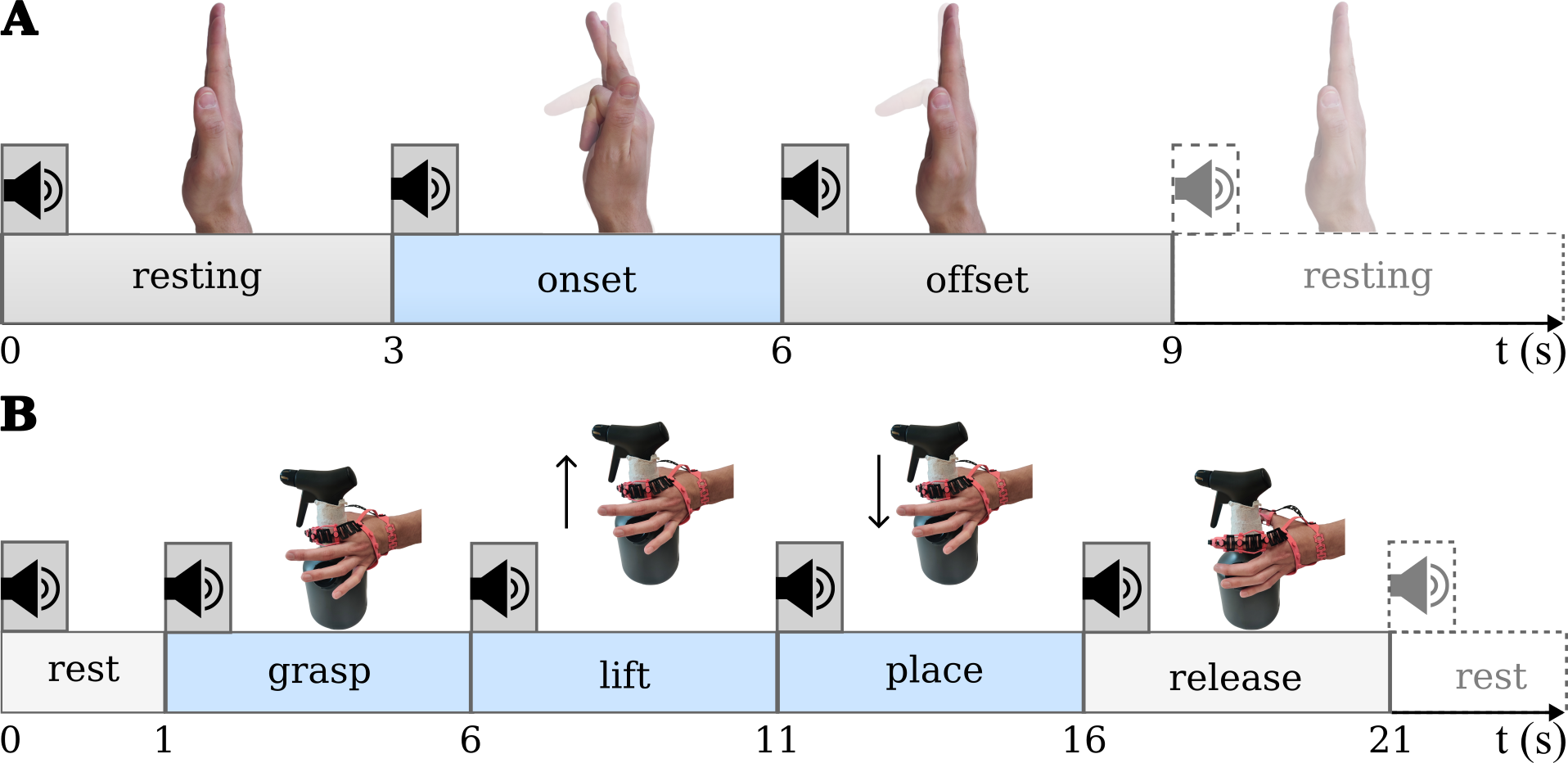}
    \caption{A visualization of the experimental protocol for one trial and its corresponding periods. The hands shows the motion pattern during each period}
    \label{fig:expSet_protocol}
\end{figure}

\textit{Index and Thumb Flexion Classification}\newline
A total of 17 volunteers participated in the data acquisition: 16 males and one female, aged 23\text{--}43 years ($28 \pm 5.5$), all right-handed and without neurological or musculoskeletal disorders affecting activities of daily living.
Each participant completed three sessions per finger motion, with each session containing 30 trials of \SI{9}{s}.
Each trial was divided into rest, onset, and offset periods, guided by voice commands at $t=0$, $t=3$, and $t=6$, respectively, as shown in Fig.~\ref{fig:expSet_protocol}\textbf{A}.
At full finger flexion, a gentle press against the palm was performed to elicit stronger muscular activation.
Participants first completed the index finger sessions, followed by the thumb sessions after a break, resulting in 90 trials per finger.\newline
\noindent\textit{Muscular Effort During Isolated Finger Flexion}\newline
To evaluate the exoskeleton during isolated finger flexion, muscular effort was recorded under three conditions: no exoskeleton, passive exoskeleton, and active exoskeleton assistance.
The protocol in Fig.~\ref{fig:expSet_protocol}\textbf{A} was adjusted to \SI{1}{s} rest, \SI{5}{s} onset, and \SI{5}{s} offset periods to allow sufficient time for the actuator to complete the finger motion.
A single participant performed ten repeated bending motions per condition while EMG activity was recorded.
The data were preprocessed as described in Subsection~\ref{sec:processing}, except that z-score normalization was replaced by maximum voluntary contraction (MVC) normalization relative to pinch lifting a \SI{1}{kg} cylinder-shaped object.\newpage
\noindent \textit{Muscular Effort During Object Grasping}\newline
The grasping experiment evaluated muscular demand during grasping and lifting of a spray bottle under two conditions: without the exoskeleton and with active exoskeleton assistance.
The same participant performed ten repetitions of grasping, lifting, placing, and releasing the object, as illustrated in Fig.~\ref{fig:expSet_protocol}\textbf{B}.
To examine the effect of object weight, the bottle was filled with water to create five loading conditions from \SI{0}{kg} to \SI{1}{kg} in \SI{0.25}{kg} increments, while the empty bottle weighed \SI{278}{g}.

\section{Results and Discussion}
\begin{figure}[H]
    \centering
    \includegraphics{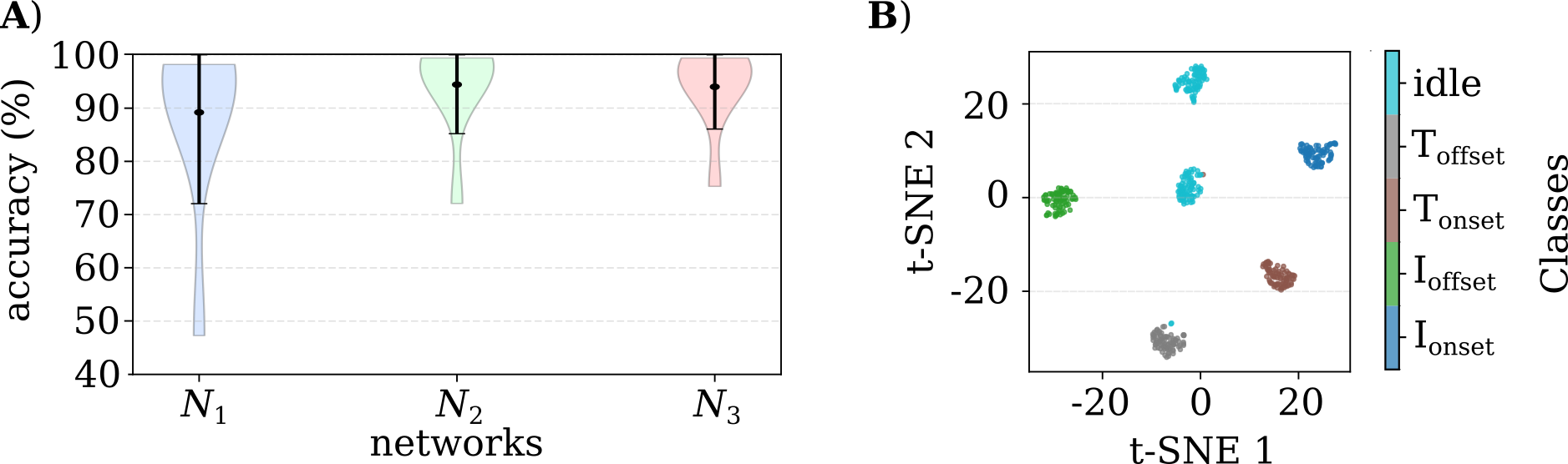}
    \caption{\textbf{A}) Performance of three models recognizing index/thumb flexion, extension, and rest (validation accuracy reflects total correct predictions across 17 participants). \textbf{B}) High\text{-}dimensional clustering of $N_2$ data across the target classes: rest (idle), thumb extension ($\mathrm{T_{offset}}$), flexion ($\mathrm{T_{onset}}$), index extension ($\mathrm{I_{offset}}$), and flexion ($\mathrm{I_{onset}}$).
    }
    \label{fig:rts_first_part}
\end{figure}

\begin{figure}[ht]
    \centering
    \includegraphics{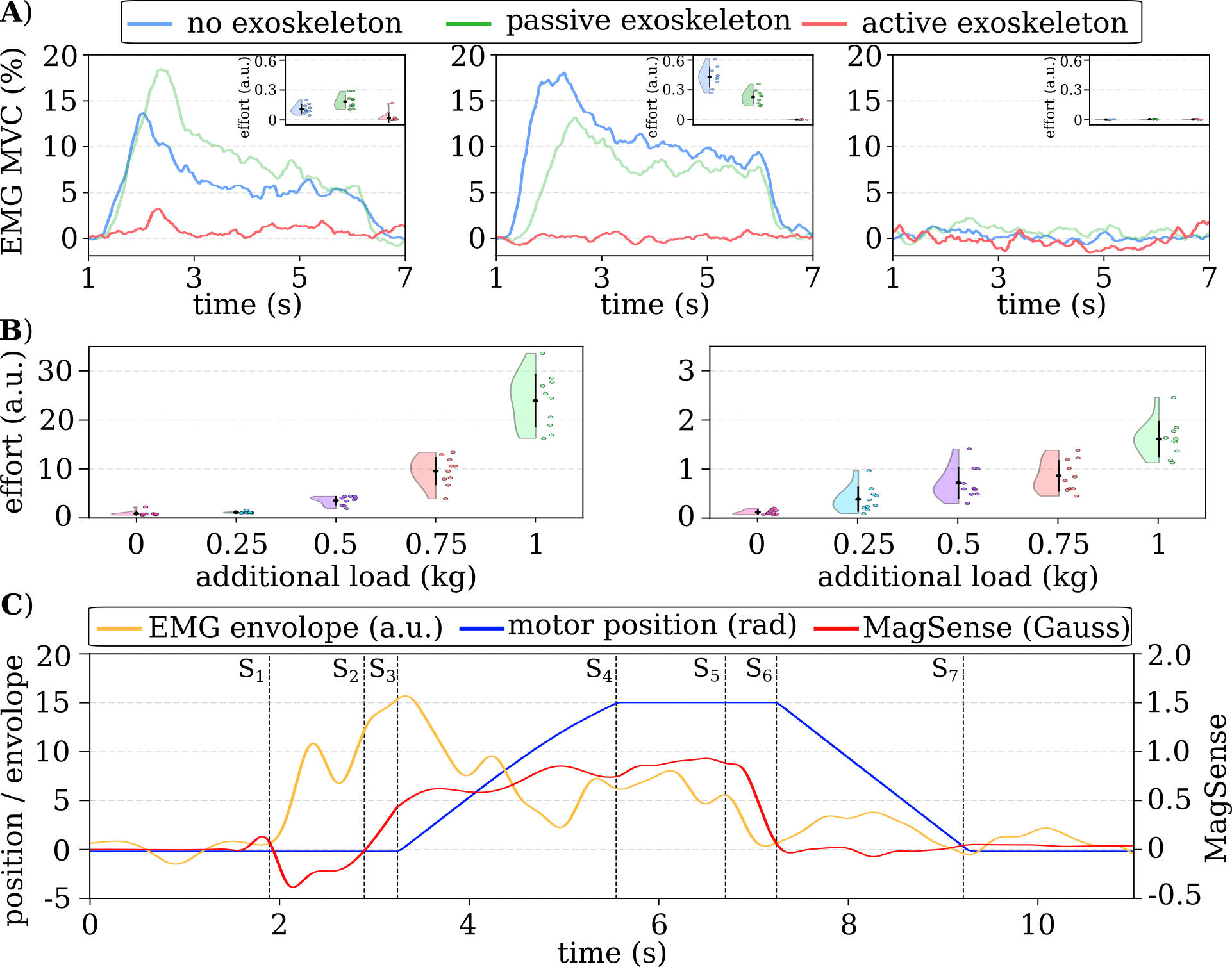}
    \caption{\textbf{A}) MVC activation for index (left), thumb flexion (middle), and pinch (right) across $n_e$, $p_e$, and $a_e$ conditions, insets show corresponding muscular effort over ten repetitions.
    \textbf{B}) Muscular effort during loaded grasping (0\text{--}1~kg), comparing unassisted and active exoskeleton conditions. \textbf{C}) Real-time grasping cycle showing EMG envelope, motor position, and magnetic field response across states $S_1\text{--}S_7$.
    }
    \label{fig:rts_second_part}
\end{figure}

Detailed insights into the exoskeleton experiments, alongside demonstrations of grasping everyday objects, are provided in Appendix~\ref{sec:appendix}. Additionally, the model architecture development is available in our GitHub repository.\newline
\textit{Index and Thumb Flexion Classification}\newline
This section evaluates the ability of the three architectures to classify index finger and thumb flexion, extension, and rest from processed EMG data, with emphasis on inter-subject robustness. K\text{--}fold accuracies in Fig.~\ref{fig:rts_first_part}\textbf{A} were compared using a Friedman test, showing a significant model effect ($\chi^2(2)=11.8$, $p<0.01$). The baseline LSTM, $N_1$, showed the largest fold-to-fold variation, indicating weaker generalization. In contrast, $N_2$ ($W \approx 0$, $p<0.01$) and $N_3$ ($W=2.0$, $p=0.02$) significantly outperformed $N_1$ in Wilcoxon signed-rank tests, suggesting that CNN-based feature extraction improves subject-invariant EMG representations. Adding attention in $N_3$ slightly reduced performance spread but did not significantly improve over $N_2$ ($W=7.0$, $p=0.38$). In the LOSO test, $N_2$ and $N_3$ both achieved $99.4\%$ accuracy, compared with $97.8\%$ for $N_1$.
\noindent Based on these results, the CNN+LSTM model was selected for real-time deployment, as the standalone LSTM showed lower accuracy and higher inter-subject variability, while attention did not significantly improve performance. Thus, $N_2$ provided the best trade-off between accuracy, simplicity, and computational efficiency.
Fig.~\ref{fig:rts_first_part}\textbf{B} shows the t-SNE embedding of the feature space learned by $N_2$. The motion classes form distinct clusters, indicating discriminative EMG representations, with only minor misclassifications mainly between idle and thumb extension.\\
\textit{Muscular Effort During Isolated Finger Flexion}\newline
Fig.~\ref{fig:rts_second_part}\textbf{A} shows muscle activity across ten repetitions for isolated index flexion (left), thumb flexion (middle), and pinch grasp (right) under three conditions: no exoskeleton ($n_e$), passive support ($p_e$), and active assistance ($a_e$).
\noindent For index flexion, the $p_e$ condition significantly increased muscular effort compared to $n_e$ ($W=8.0$, $p=0.048$), while $a_e$ reduced effort compared to both cases ($W=1.0$, $p<0.01$).
Thumb flexion showed significant differences across all conditions ($W\approx0$, $p<0.01$).
Specifically, comparing $p_e$ to $n_e$ revealed that index flexion slightly increased muscular demand due to exoskeleton resistance, whereas thumb flexion reduced it likely because the exoskeleton restricted thumb motion, lowering recorded muscle activity.
Nevertheless, $a_e$ significantly reduced muscular demand in both cases.
Conversely, the pinch grasp showed no significant differences between conditions ($\chi^2(2)=1.3$, $p=0.51$).
Because unloaded pinch motion requires minimal effort, the signal merely fluctuated around the baseline noise floor, prompting an analysis of the loaded pinch grasp to evaluate exoskeleton assistance during more demanding tasks.

\noindent\textit{Muscular Effort During Object Grasping}\newline
Fig.~\ref{fig:rts_second_part}\textbf{B} illustrates the EMG-derived muscular effort during repeated grasping and lifting of the spray bottle. For $n_e$ and $a_e$ conditions, muscular effort increased with additional loads, indicating higher demand during heavier lifting.
Without exoskeleton assistance, muscular effort increased
exponentially from the unloaded to the \SI{1}{kg} condition,
where all except the two lightest cases showed a significant
difference ($p < 0.05$).
On the contrary, active exoskeleton assistance consistently reduced muscular effort across all loads.
Even during the \SI{1}{kg} lifting task, the active exoskeleton condition produced significantly lower muscular demand compared to the corresponding no\text{--}exoskeleton \SI{0.25}{kg} condition ($W\approx0$, $p=0.002$).
In general, the muscular effort was reduced by a factor of ten when assisted by the exoskeleton.
The results further suggest that muscular effort under active assistance increased more gradually with additional load. Several neighboring loads between \SI{0.25}{kg} and \SI{0.75}{kg} did not differ significantly, indicating that the exoskeleton partially compensates for the lifting demand until exceeding \SI{1}{kg}, where users must contribute their own force.
Overall, the experiment demonstrated that active exoskeleton assistance reduced muscular effort during functional grasping and lifting across varying loads, with a reduction of $92.6\%$ observed between the two conditions at the highest load.\newline
\noindent \textit{Real-time control of exoskeleton}\newline
The $N_2$ model was tested in real time to control the exoskeleton during active assistance. The participant was not included in the training data, showing that the model could work with an unseen user. During the demonstration, the participant initiated a pinch motion, which was detected by the model and used to activate the exoskeleton for grasping everyday objects.
Fig.~\ref{fig:rts_second_part}\textbf{C} illustrates the signal behavior of the EMG activity, motor position, and MagSense during grasping of a water bottle. Initially, the participant remains in an idle state until $S_1$, where a voluntary pinch motion is initiated, causing an increase in EMG activity. At $S_2$, MagSense detects object contact, while the increasing EMG activity allows the model to recognize the need for assistance. At $S_3$, the model predicts motion intention and commands the ESP controller to actuate the motors. The motor position subsequently increases to approximately \SI{15}{rad} and stops once the MagSense reaches a predefined threshold, indicating that the object has been fully grasped. The grasp is maintained until a voluntary release is initiated at $S_5$, resulting in a decrease in both EMG activity and magnetic field measurements. At $S_6$, the model recognizes the release pattern and assists the motors in releasing their tension. Finally, at $S_7$, the system returns to the resting state.

\section{Conclusion}
This work presented SoftPINCH, a soft wearable hand exoskeleton for assisting thumb and index finger movement during pinch grasping. The system combines EMG-based intention detection, tendon-driven assistance, and fingertip contact sensing. Among the tested control models, CNN+LSTM was selected for real-time use because it achieved high accuracy while remaining simpler than the attention-based model. The system reached 99.4\% test accuracy for movement recognition. Active assistance reduced muscle effort during finger flexion and object grasping, with a 92.6\% reduction during the heaviest lifting task.
Overall, these results show that SoftPINCH can provide accurate EMG-based control and meaningful assistance for pinch grasping. Future work will focus on testing the system with more users, including individuals with hand impairments, and evaluating its performance in daily tasks involving repeated grasping, holding, and lifting. In addition, the exoskeleton controller will be extended to adapt the applied assistance force and actuation speed based on the user’s EMG activity.

\section*{Appendix}\label{sec:appendix}
Supplementary video: \url{https://youtu.be/dmQ18kg8WZo}\\
Source code and supplementary implementation details are available at: \\\url{https://github.com/SDUSoftRobotics/SoftPINCH}

\bibliographystyle{splncs04}
\bibliography{refs}

\end{document}